%% file: main.tex
\pdfoutput=1

\documentclass[11pt, dvipsnames]{article}

\usepackage[]{main}
\usepackage[shortlabels]{enumitem}
\usepackage{times}
\usepackage{color,soul} 
\usepackage{booktabs}
\usepackage{hyperref}
\usepackage{latexsym}
\usepackage{hyperref}
\usepackage{booktabs}
\usepackage{multirow}
\usepackage[LAE,T1]{fontenc}
\usepackage{arabtex}
\usepackage{utf8} 
\usepackage[utf8]{inputenc} 
\usepackage[LAE,T1]{fontenc}
\usepackage[arabic,french,english]{babel}
\usepackage[raggedrightboxes]{ragged2e}
\usepackage{todonotes}
\usepackage{array}
\usepackage{lscape}
\usepackage{natbib}
\usepackage{multirow}
\usepackage{float}
\usepackage{makecell}
\usepackage{xcolor, colortbl}
\usepackage{inconsolata}
\usepackage{arydshln}
\usepackage{array}

\newcolumntype{H}{>{\setbox0=\hbox\bgroup}c<{\egroup}@{}}
\usepackage{cancel}
\usepackage{ulem,xpatch}

\usepackage[T1]{fontenc}

\usepackage[utf8]{inputenc}

\usepackage{microtype}
\usepackage{rotating}

\usepackage{inconsolata}

\usepackage{graphicx}
\usepackage{subcaption}
\usepackage{multirow}
\usepackage{tabularx}
\usepackage{framed}
\newcommand\blfootnote[1]{%
  \begingroup
  \renewcommand\thefootnote{}\footnote{#1}%
  \addtocounter{footnote}{-1}%
  \endgroup
}

\usepackage{dblfloatfix}

\newcommand{\toolname}{VoxArabica}
\newcommand{\system}{VoxArabica}

\title{\includegraphics[scale=0.08]{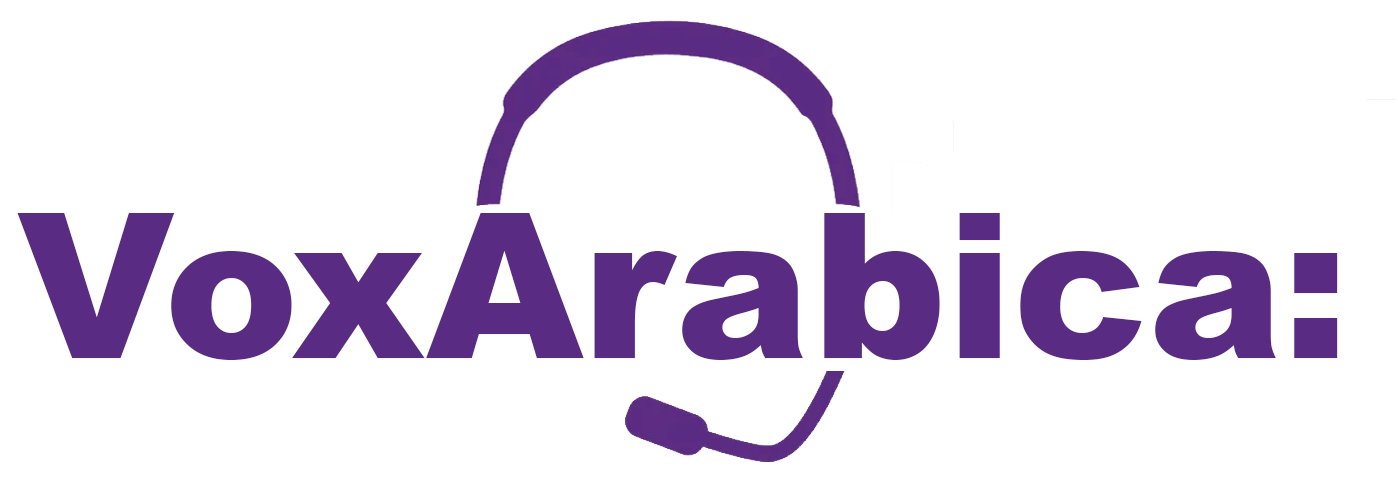} \\ A Robust Dialect-Aware Arabic Speech Recognition System}

\author{\normalsize Abdul Waheed$^{\lambda,\star}$~Bashar Talafha
$^{\xi,\star}$~Peter Sullivan$^{\xi,\star}~$ \\
\normalsize \bf AbdelRahim Elmadany$^{\xi}$~Muhammad Abdul-Mageed$^{\xi,\lambda}$ \\
\normalsize $^{\xi}$ Deep Learning \& Natural Language Processing Group,
  The University of British Columbia\\\normalsize  $^{\lambda}$Department of Natural Language Processing \& Department of Machine Learning, MBZUAI\\ %
  \texttt{\normalsize muhammad.mageed@ubc.ca}}

\begin{document}
\setcode{utf8}
\maketitle
\section*{~~~~~~~~~~~~~~~~~~~~~~~~~~~~~Abstract}
Arabic is a broad language with many varieties and dialects spoken by $\sim450$ millions all around the world. Due to the linguistic diversity and variations, it is challenging to build a robust and generalized ASR system for Arabic. In this work, we address this gap by developing and demoing a system, dubbed \toolname, for dialect identification (DID) as well as automatic speech recognition (ASR) of Arabic. We train a wide range of models such as HuBERT (DID), Whisper, and XLS-R (ASR) in a supervised setting for Arabic DID and ASR tasks. Our DID models are trained to identify $17$ different dialects in addition to MSA. We finetune our ASR models on MSA, Egyptian, Moroccan, and mixed data. Additionally, for the remaining dialects in ASR, we provide the option to choose various models such as Whisper and MMS in a zero-shot setting. We integrate these models into a single web interface with diverse features such as audio recording, file upload, model selection, and the option to raise flags for incorrect outputs. Overall, we believe \toolname~ will be useful for a wide range of audiences concerned with Arabic research. Our system is currently running at~\url{https://cdce-206-12-100-168.ngrok.io/ }.


\input{sections/1_introduction}
\input{sections/2_lit_review}

\input{sections/3_methodology}

\input{sections/4_walkthrough}

\input{sections/6_conclusion}
\input{sections/7_limitations}
\input{sections/8_ethics_statement}
\input{ack}
\phantomsection  

\normalem
\bibliography{custom}
\bibliographystyle{acl_natbib}
\clearpage 
\appendix

\input{sections/9_appendix}
\end{document}

%% file: sections/1_introduction.tex
\section{Introduction}\label{sec:introduction}
 ~\blfootnote{ $^{\star}${Equal contributions}}
The Arabic language, with its diverse regional dialects, represents a unique linguistic spectrum with varying degrees of overlap between the different varieties at all linguistic levels (e.g., phonetic, syntactic, and semantic). In addition to Modern Standard Arabic (MSA), which is primarily used in education, pan-Arab media, and government, there are many local dialects and varieties that are sometimes categorized at regional~
\cite{zaidan2014arabic, elfardy2013sentence, elaraby2018deep}, country~\cite{bouamor2018madar, abdul2020nadi, abdul2021nadi, abdul2022nadi}, or even province levels~\cite{abdul2020toward}. 
Historically, this wide and rich variation between different Arabic varieties has posed a significant challenge for automatic speech recognition (ASR)~\cite{talafha2023n, alsayadi2022deep, ali2020multi}. The main focus has largely been on the recognition of MSA with very little-to-no focus on its dialects and varieties~\cite{app12178898, hussein2022arabic, Ali2014AdvancesID}. 
As such, ASR systems have conventionally been built either for MSA or individual dialects, thereby restricting their versatility and adaptability. However, the multifaceted nature of Arabic demands a robust ASR system that caters for its diverse dialects and varieties. In this work, we fill this research gap by introducing and demoing an ASR system integrated with a dialect identification model, dubbed~\textit{\system}. 

\system~ is an end-to-end dialect-aware ASR system with dual functionality: (i) it offers a supervised dialect identification model followed by (ii) a finetuned Whisper Arabic ASR model covering multiple dialects. The dialect identification model works by assigning a country-level dialect, as well as MSA, from a set of $18$ labels from input speech. This then allows the appropriate ASR model to fire. Contrary to traditional methodologies that separate dialect identification and speech recognition as two completely different tasks, our proposed pipeline integrates the two components effectively utilizing dialectal information for improved speech recognition. Such an integration not only improves the ASR output, but also establishes a framework aligned with the linguistic diversities inherent to Arabic as well. Concretely, our contributions can be summarized as follows:
\begin{itemize}
    \item We introduce and demo our end-to-end~\system~system, which integrates dialect identification with state-of-the-art Arabic ASR.
    \item Our demo is based on a user-friendly web interface characterized with rich functionalities such as audio uploading, audio recording, and user feedback options. 
\end{itemize}

The rest of the paper is organized as follows: In Section ~\ref{sec:lit_review}, we overview related works. Section~\ref{sec:methods} introduces our methods. Section ~\ref{sec:walkthrough} offers a walkthrough of our demo. We conclude in Section~\ref{sec:conclusion}.

%% file: sections/2_lit_review.tex
\section{Literature Review}\label{sec:lit_review}

\noindent \textbf{Arabic ASR.} Recent ASR research has focused on end-to-end (E2E) methods such as in Whisper~\cite{radford2022robust} and the Universal Speech Model~\cite{zhang2023google}.
Such E2E deep learning models have significantly elevated ASR performance by allowing learning directly from the audio waveform, bypassing the need for intermediate feature extraction layers~\cite{wang2019overview, radford2022robust}. Whisper is particularly noteworthy for its multitask training approach, incorporating ASR, voice activity detection, language identification, and speech translation. It has achieved state-of-the-art performance on multiple benchmark datasets such as Librispeech~\cite{panayotov2015librispeech} and TEDLIUM~\cite{rousseau2012ted}. However, its resilience to adversarial noise has been questioned~\cite{olivier2022there}.

For Arabic ASR specifically, the first E2E model was introduced using recurrent neural networks coupled with Connectionist Temporal Classification (CTC)~\cite{ahmed2019end}. Subsequent works have built upon this foundation, including the development of transformer-based models that excel in both MSA and dialects~\cite{belinkov2019analyzing,hussein2022arabic}. One challenge for E2E ASR models is the substantial requirement for labeled data, particularly for languages with fewer resources such as varieties of Arabic. To address this, self-supervised and semi-supervised learning approaches are gaining traction. These models, such as Wav2vec2.0 and XLS-R, initially learn useful representations from large amounts of unlabeled or weakly labeled data and can later be finetuned for specific tasks~\cite{baevski2020Wav2vec,babu2021xls}. 
W2v-BERT, another self-supervised model, employs contrastive learning and masked language modeling. It has been adapted for Arabic ASR by finetuning on the FLEURS dataset, which represents dialect-accented standard Arabic spoken by Egyptians~\cite{chung2021w2v,conneau2023fleurs}. Unlike Whisper, both Wav2vec2.0 and w2v-BERT necessitate a finetuning stage for effective decoding. 


\noindent \textbf{Arabic DID.}
Arabic DID has been the subject of a number of studies through recent years, enhanced by collection of spoken Arabic DID corpora such as ADI5~\cite{ali2017speech} and ADI17~\cite{shon2020adi17}. And advances in model architecture have mirrored changes in the larger LID research community, from i-vector~\cite{dehak2010front} based approaches~\cite{ali2017speech} towards deep learning based approaches: x-vectors~\cite{snyder2018x,shon2020adi17}, end-to-end classification using deep neural networks~\cite{ali2019mgb,cai2018insights}, and transfer learning~\cite{sullivan23_interspeech}. 

\noindent \textbf{ASR and DID.} Combining ASR and DID in a single pipeline remains fairly novel for Arabic. Recent works in this space has employed only limited corpora~\cite{lounnas2020cliasr}, or used ASR transcripts only to improve DID~\cite{malmasi-zampieri-2017-arabic}. 
Closest to our demonstrated system in this work is FarSpeech~\cite{eldesouki2019farspeech}, since it combines ASR and DID. However, FarSpeech is confined to coarse-grain DID and only supports MSA for ASR. In addition, compared to FarSpeech, our models are \textit{modular} in that it allows users to run either or both ASR or DID, depending on their needs.

\input{figures/main-figures}

%% file: figures/main-figures.tex

    


\begin{figure*}[h!]
\vspace{-1.3cm}
\includegraphics[width=\textwidth]{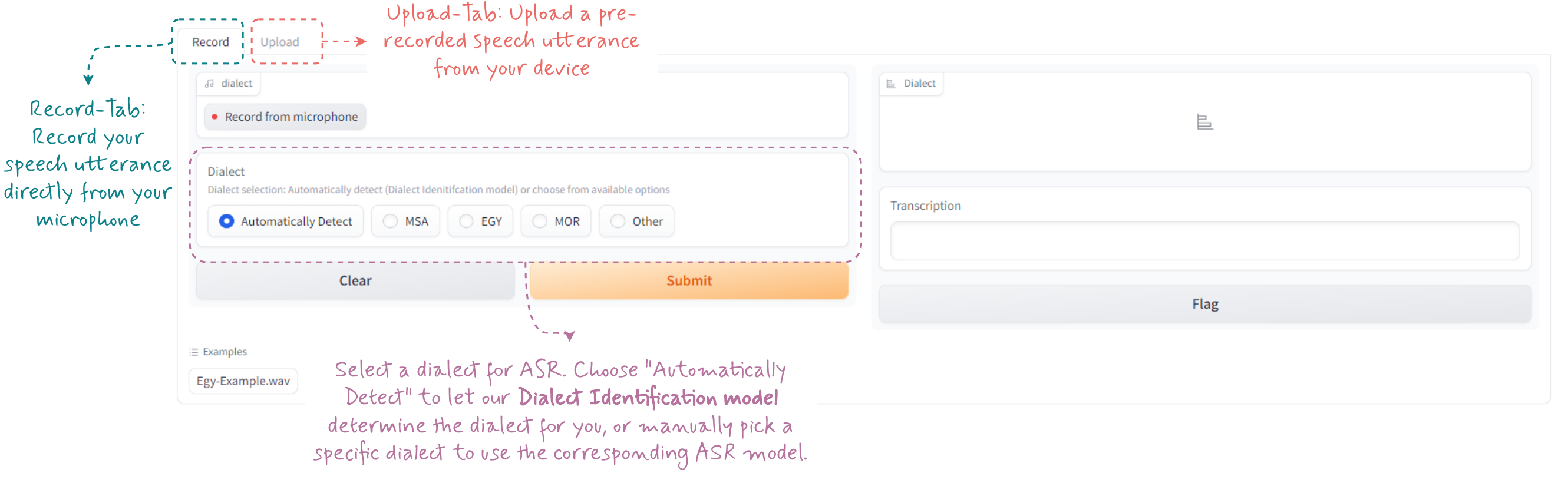}
\caption{\label{upload-record-feature}
Users have the \textbf{option to either upload files or directly record their audio}. Additionally, the dialect can be automatically detected or manually selected for a specific ASR model.}
\label{fig1}
\end{figure*}


\begin{figure*}[h!]
\includegraphics[width=\textwidth]{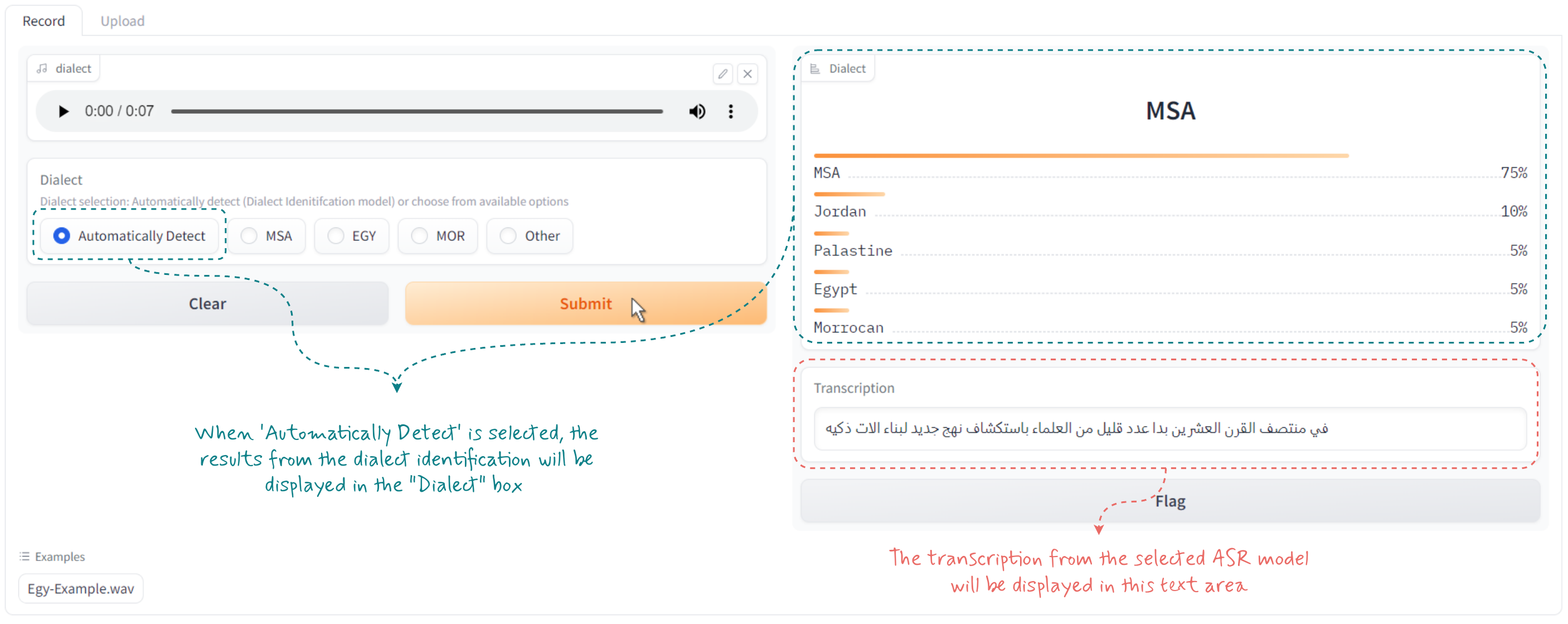}
\caption{\label{did}
For \textbf{automatic dialect detection}, likelihood percentages determine the ASR model choice, with transcriptions displayed in the Transcription text area.}
\label{fig3}
\end{figure*}


\begin{figure*}[h!]
\centering
 \includegraphics[width=0.8\textwidth]{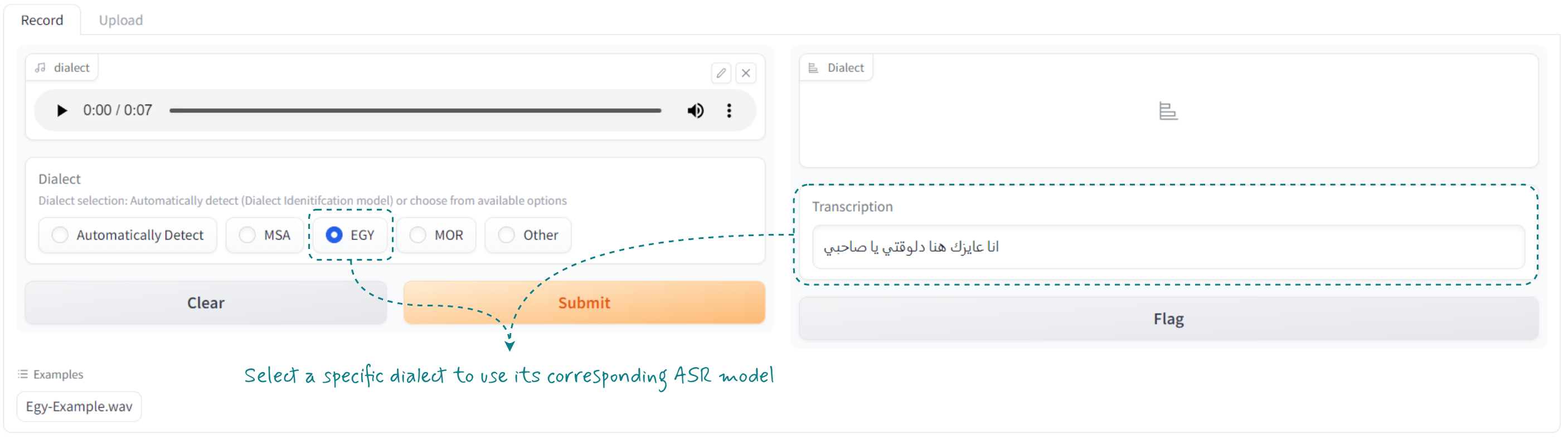}
\includegraphics[width=0.8\textwidth]{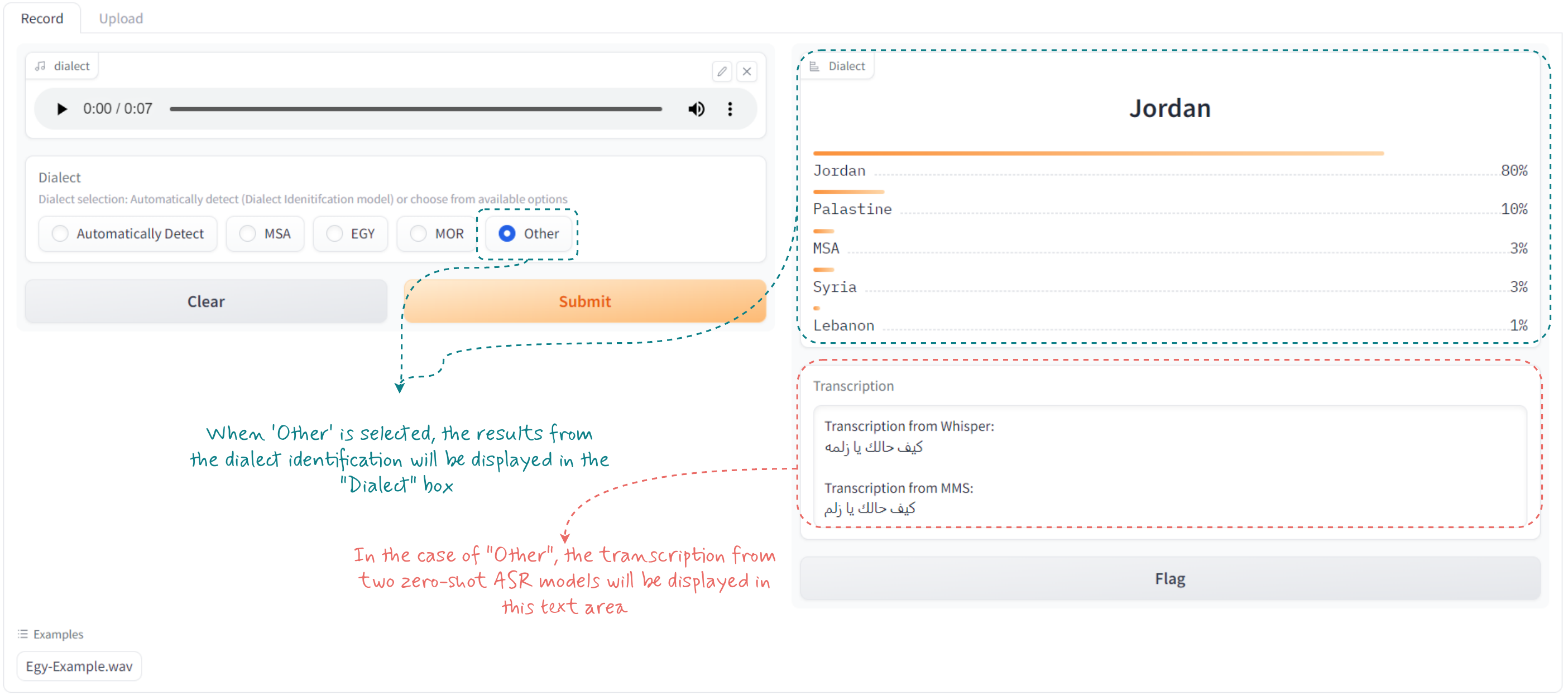}
\caption{\label{asr}
When a specific dialect is manually selected, its associated \textbf{ASR model generates the transcription}. When recording in an unlisted dialect, select "Other". The dialect identification model will then detect the dialect, and both Whisper and MMS zero-shot models will produce the transcription.}
\label{fig4}
\end{figure*}

%% file: sections/3_methodology.tex
\section{Models}\label{sec:methods}

\subsection{DID Models}
Our DID model is a transfer learning approach: finetuning HuBERT~\cite{hsu2021hubert} on ADI-17~\cite{shon2020adi17} and the MSA portions of ADI-5~\cite{ali2017speech} and MGB-2~\cite{ali2016mgb}. We utilize only the MSA portions of ADI-5 due to the ambiguity of going from coarse-grain to fine-grain labels. Dialectal varieties covered in our model are \textit{MSA, Algerian, Egyptian, Iraqi, Jordanian, Saudi, Kuwaiti, Lebanese, Libyan, Mauritanian, Moroccon, Omani, Palestinian, Qatari, Sudanese, Syrian, Emirati}, and \textit{Yemeni}. 

\textbf{Training Details.} Our finetuning procedure entailed performing a random search for training hyperparameters validated using the ADI-17 development set. A detailed overview of the hyperparameters searched can be found in Table \ref{tab:hyperparams}
\input{tables/did_hyperparam}. 
We train using AdamW as optimizer, with a certain number of initial steps, \textit{Freeze Steps}, where the original model is not updated and only the newly initialized classification layers change. After thawing, we also experiment with keeping some of the earlier layers of the model frozen. We indicate the earliest layer that gets thawed as \textit{Thaw Depth}. We also experiment with LayerNorm and Attention finetuning~\cite{li2020multilingual}, but our final model performed better without it.

\subsection{ASR Models}
We train a wide range of ASR models on a list of benchmark Arabic speech datasets. Our models include two versions of Whisper ~\cite{radford2022robust}, \textit{large-v2} and \textit{small}. We also finetune XLS-R~\cite{babu22interspeech} for the ASR task. For MSA, we train our models on three versions of \textit{common voice}~\cite{ardila2019common} datasets \texttt{6.1, 9.0}, and \texttt{11.0}. We note that~\citet{talafha2023n} show that Whisper \textit{large-v2} outperforms its smaller variant as well as XLS-R trained on the same dataset. For Morrocan, Egyptian, and MSA, we fully finetune models on MGB2, MGB3, MGB5~\cite{ali2016mgb, ali2017speech, ali2019mgb}. We also train ASR models on FLEURS~\cite{conneau2023fleurs}, which is accented Egyptian speech data. 

\noindent\textbf{Text Preprocessing.}\label{sec:text-preprocessing-appendix}
The datasets we employ exhibit various inconsistencies. For instance, within CV6.1, the utterance \<فَقَالَ لَهُمْ> "faqaAla lahumo" is fully diacritic, whereas the utterance \<فإذا النجوم طمست> "f$<$*A Alnjwm Tmst" lacks diacritic annotations, despite both originating from the Quran. Consequently, we adopt the normalization approach from \cite{chowdhury21interspeech, talafha2023n}, which involves: (a) discarding all punctuation marks excluding the \% and @ symbols; (b) eliminating diacritics, Hamzas, and Maddas; and (c) converting eastern Arabic numerals into their western counterparts (e.g., 29 remains 29). Given that this study does not address code-switching, all Latin alphabet is excluded.\\
\textbf{Training Details.} 
Before training, we apply pre-processing steps as mentioned above on the text. We train all of our models using AdamW optimizer~\cite{loshchilov2018decoupled} with a learning rate of $1e-5$, 500 warmup steps, and no weight decay. To prevent the model from severely overfitting, we employ early stopping with patience at $5$. We use Huggingface trainer~\footnote{\url{https://huggingface.co/docs/transformers/main_classes/trainer}} with deepspeed ZeRO~\cite{DBLP:journals/corr/abs-1910-02054} stage-2 to parallelize our training across 8xA100 (40G) GPUs.

In our demo, we also allow users to utilize both Whisper and MMS~\cite{pratap2023scaling} in the zero-shot setting. 


\input{tables/examples}

%% file: tables/did_hyperparam.tex
\begin{table}
\caption[Dialect ID hyperparameters]{
An overview of the search space of the hyperparmeter tuning for the DID model as well as optimal configuration found during the (n=30) random search. The batch size formula ensures our V100 GPUs were fully utilized during training, with a target of $75$ seconds of audio regardless of the sampling duration. All values are picked from uniform distributions except for the learning rate, which was picked from a log uniform distribution.  
}
\label{tab:hyperparams}
\centering
\resizebox{0.95\columnwidth}{!}{%
\begin{tabular}{lcl} \toprule 
 \textbf{}  & \textbf{Range} & \textbf{Conf.}  \\
 \midrule
     Batch Size  & $4 \cdot \lfloor\frac{75}{Duration} \rfloor$  & $16$ \\ 
 Freeze Steps  & $[0,1000]$ & $192$ \\

 Learning Rate  & $[1 \cdot  10^{-5},1 \cdot 10 ^{-2}]$ &$6 \cdot 10^{-4}$  \\

 Max Steps & $[20k,40k]$ & $29225$ \\ 
  Duration  & $[4,18]$ seconds & $4.69$ \\ 

 Thaw Depth  & $[0,23]$ & $3$\\ 

 \bottomrule 
\end{tabular}%
}
\end{table}

%% file: tables/examples.tex
\begin{table}[ht]
\centering
\footnotesize 
\renewcommand{\arraystretch}{0.5}   
\resizebox{1.0\linewidth}{!}{%
\begin{tabular}{lp{7cm}}
\toprule
\cellcolor{LimeGreen!30}Ref (EGY) & \cellcolor{LimeGreen!30} \RaggedRight{\begin{arabtext} \small   مساء الخير اهلا ومرحبا بيكم في حلقة جديدة من برنامج بوضوح اي واحد نفسه في ثانية يطلعها قدام الكاميرا \end{arabtext}} \\ \hdashline 

 \cellcolor{LimeGreen!10} Whisper (0-shot) & \cellcolor{LimeGreen!10} \RaggedRight{\begin{arabtext}
    \small بسعي الخير أهلا ومرحبا بكم في حلقة جديدة من برنامج بوضوح أي واحد نفسه في ثانية يطلعها قدام الكاميرا
\end{arabtext}} \\ 
\cellcolor{Plum!10}MMS (0-shot) & \cellcolor{Plum!10}\RaggedRight{\begin{arabtext}
    \small بساء لخير أهلن مرحباً بكم فحلى أجديدة من برنامج بوضوح أي واحد نفسفسنية يطلعها ودمك كمرة
\end{arabtext}} \\ \hdashline

 Whisper (MSA) & \RaggedRight{\begin{arabtext}
    \small بسعر الخير اهلا ومرحبا بكم في حلقة جديدة من برنامج بوضوح اي واحد نفسه سامي لا يطلعها قدم الكاميرا
\end{arabtext}} \\

 \cellcolor{LimeGreen!10} Whisper(EGY) & \cellcolor{LimeGreen!10} \RaggedRight{\begin{arabtext}
    \small مساء الخير اهلا ومرحبا بيكم في حلقة جديدة من برنامج بوضوح اي واحد نفسه في ثانية يطلعها قدام الكاميرا
\end{arabtext}} \\

 \cellcolor{Plum!10} Whisper(MOR)  & \cellcolor{Plum!10} \RaggedRight{\begin{arabtext}
    \small مسايا الخير اهلا ومرحبا بيكم في حلقة جديدة من برنامج بوضوح اي واحد نفسو فسنية لي يطلعها قدام الكاميرا
\end{arabtext}} \\ \hdashline
XLS-R(MSA) & \RaggedRight{\begin{arabtext}
    \small مساء الخير اهلا ومرحبا بكم في حالة جديدة من برنامج بوضوح اي واحد نفسه ثانية يطلعها قدام الكاميرا
\end{arabtext}} \\

\bottomrule
\end{tabular}
}
\caption{\label{tab:examples}
    Example outputs produced by \system~when input audio is Egyptian dialect. 
}
\end{table}

%% file: sections/4_walkthrough.tex
\section{Walkthrough}\label{sec:walkthrough}

Our demo consists of a web interface with versatile functionality. It allows users to interact with the system in multiple ways, depending on their needs. 

\noindent \textbf{User audio input.} Users can either record their own audio through a microphone or upload a prerecorded file. In both cases, we allow different formats such as \texttt{.wav}, \texttt{.mp3}, or \texttt{.flac}, across various audio sampling rates (e.g., \texttt{16khz} or \texttt{48khz}).
Figure~\ref{upload-record-feature} demonstrates the different options available to the user upon interacting with~\system.

\noindent \textbf{Model selection.} Users can choose to select an Arabic variety for transcription, or have it automatically detected using our $18$-way DID system. We demonstrate this in Figure~\ref{did}. Once the variety is detected, the corresponding ASR model will perform transcription and both DID transcription results will be presented on the interface (as shown in Figure \ref{asr}). We offer various models: two for the EGY and MOR, respectively; two for MSA; and two generic models that can be used for any variety. We list all models in Table~\ref{tab:models}. In cases where predicted/selected variety is not covered by our ASR models, we fall back to our generic models (i.e., both Whisper zero-shot and MMS zero-shot).  

\begin{table}[]
\centering
\footnotesize 
\renewcommand{\arraystretch}{0.9}   
\resizebox{0.99\linewidth}{!}{%
\begin{tabular}{lccl}
\toprule
\textbf{Model name} & \textbf{Dialect(s)} & \textbf{Dataset} & \textbf{Architecture} \\ \midrule
Whisper MSA         & MSA    & CV (6.1, 9.0, 11.0)             & Whisper              \\
XLS-R               & MSA   & CV (6.1, 9.0, 11.0)               & Wav2vec 2.0           \\
Whisper Morroco     & MOR   & MGB5        & Whisper               \\
Whisper Egypt       & EGY   & MGB3         & Whisper               \\
Whisper Zero-shot   & -     & -        & Whisper               \\
MMS                 & -     & -          & Wav2vec 2.0           \\ \bottomrule
\end{tabular}
}
\caption{The utilized ASR models, their associated dialects, and respective architectures, and dataset used to train each model. Models marked with a dash are generic and not specific to a particular dialect.}
\label{tab:models}
\end{table}

\noindent \textbf{User feedback.} We also provide an option for users to submit \textit{anonymous} feedback about the produced output by raising a flag. We use this information to collect high quality silver labels and discard examples where a flag is raised for incorrect outputs. It is important to note that we do not collect any external user data for any purpose, thus ensuring user privacy. 

\noindent \textbf{System output.}
Our system conveniently outputs both predicted Arabic variety and transcription across two panels as shown in Figure~\ref{asr}. For predicted variety, we show users all top five predictions along with model confidence for each of them. 
We provide outputs produced by our models in \toolname~when the reference input is Egyptian dialect in Table ~\ref{tab:examples}. We also present additional examples in Appendix ~\ref{appendix-sec:examples}, Table~\ref{tab:examples-appendix}.


%% file: sections/6_conclusion.tex
\section{Conclusion}\label{sec:conclusion}

We present a demonstration of combined DID and ASR pipeline to illustrate the potential for these systems to improve the usability of dialectal Arabic speech technologies. We report example outputs produced by our system for multiple dialects showcasing the effectiveness of integrated DID and ASR pipelines. We believe that our demo will advance the research to build a robust and generalized Arabic ASR system for a wide range of varieties and dialects and will enable a more holistic assessment of the strengths and weaknesses of these methods. For future work, we intend to add models for more dialects and varieties particularly those which are low resource.

%% file: sections/7_limitations.tex
\section{Limitations}\label{sec:limitations}

Audio classification tasks can be susceptible to out-of-domain performance degradation, which may impact real world performance. Similarly, studies on the interpretability of DID models have shown internal encoding of non-linguistic factors such as gender and channel~\cite{chowdhury2020does}, which may impart bias to the models. Ensuring training corpora contain a diverse balance of speaker gender, recording conditions, as well as full coverage of the different styles of language is an ongoing challenge. We hope that by creating an online demonstration, these limitations can be further explored.

%% file: sections/8_ethics_statement.tex
\section{Ethics Statement}\label{sec:ethics_statement}

\noindent \textbf{Intended use.} We build a robust dialect identification and speech recognition system for multiple Arabic dialects as well as MSA. We showcase the capability of our system in the demo. We believe that our work will guide a new direction of research to develop a robust and generalized speech recognition system for Arabic. Through our demo, we integrate DID with ASR system which support multiple dialects.

\noindent \textbf{Potential misuse and bias.}
Since our data is limited to a few dialects involved in finetuning DID and ASR systems, we do not expect our models to generalize all varieties and dialects of Arabic that are not supported by our models.

%% file: ack.tex
\section*{Acknowledgments}\label{sec:acknow}
We gratefully acknowledge support from Canada Research Chairs (CRC), the Natural Sciences and Engineering Research Council of Canada (NSERC; RGPIN-2018-04267), the Social Sciences and Humanities Research Council of Canada (SSHRC; 435-2018-0576; 895-2020-1004; 895-2021-1008), Canadian Foundation for Innovation (CFI; 37771), Digital Research Alliance of Canada,\footnote{\href{https://alliancecan.ca}{https://alliancecan.ca}} and UBC ARC-Sockeye.\footnote{\href{https://arc.ubc.ca/ubc-arc-sockeye}{https://arc.ubc.ca/ubc-arc-sockeye}}


%% file: sections/9_appendix.tex
\section*{Appendix}\label{sec:appendix}
\section{Example Outputs}\label{appendix-sec:examples}
See Table \ref{tab:examples-appendix} on the following page.
\input{tables/examples-appendix}

%% file: tables/examples-appendix.tex
\begin{table*}[!ht]
\centering
\footnotesize 
\renewcommand{\arraystretch}{0.9}   
\resizebox{0.99\linewidth}{!}{%
\begin{tabular}{p{4cm}p{18cm}}
\toprule
Ref (MSA) & \RaggedRight{\begin{arabtext}
    \small
    يؤثر التدخين بشكل سلبي في جسم الانسان حيث ينتج عنه العديد من الاثار السلبية المؤذية للفرد وقد تؤدي بعضها الى مضاعفات تهدد الحياة

\end{arabtext}} \\ \hdashline

MMS   &  \RaggedRight{\begin{arabtext}
    \small يؤثر التدخين بشكل سلبي هي جسم الإنسان حيث ينتجعنه العديد من الآثار السلبية المؤذيد الفرد وقد تؤدي بعضها إلى مضاعفات تهدد الحياة
\end{arabtext}} \\

Whisper(0-shot)& \RaggedRight{\begin{arabtext}
    \small يؤثر التدخين بشكل سلبي في جسم الانسان حيث ينتج عنه العديد من الاثار السلبية المؤذية للفرد وقد تؤدي بعضها الى مضاعفات تهدد الحياة
\end{arabtext}} \\ \hdashline

Whisper(MSA)& \RaggedRight{\begin{arabtext}
    \small 
  يؤثر التدخين بشكل سلبي في جسم الإنسان حيث ينتج عنه العديد من الآثار السلبية المؤذية للفرد وقد تؤدي بعضها إلى مضاعفات تهدد الحياة
\end{arabtext}} \\

Whisper(MOR) & \RaggedRight{\begin{arabtext}
    \small    يؤثر التدخين بشكل سلبي في جسم الانسان حيت ينتج عنه العديد من الاثار السلبية المؤذية دالفرق وقد تؤدي بعضها الى مضعفات تهدد الحياة
\end{arabtext}} \\

Whisper(EGY)& \RaggedRight{\begin{arabtext}
    \small يؤثر التدخين بشكل سلبي في جسم الانسان حيث ينتج عنه العديد من الاثار السلبية المؤذية للفرد وقد تؤدي بعضها الى مضاعفات تهدد الحياة

\end{arabtext}} \\
\midrule

Ref (JOR - Other) & \RaggedRight{\begin{arabtext} \small يا زلمة كيف حالك؟ شو أخبارك؟ وين هالغيبة؟ زمان عنك، ليش ما بتبين؟ \end{arabtext}} \\ 
\hdashline 
MMS (0-shot) & \RaggedRight{\begin{arabtext} \small يعزل كاف حلكشو أخبارك وانه الغاب زمان عنك لاش ما بتبين \end{arabtext}} \\ 

Whisper (0-shot) & \RaggedRight{\begin{arabtext} \smallيا زلمة كيف حالك؟ شو أخبارك؟ وين هالغابة؟ زمان عنك، ليش ما بتبين؟\end{arabtext}} \\ \hdashline
Whisper (MSA) & \RaggedRight{\begin{arabtext} \small يا زلم كيف حالك شو اخبارك وانها الغيبة زمان عنك ليش ما بتبين \end{arabtext}} \\ 
Whisper (MOR) & \RaggedRight{\begin{arabtext} \small يا زلمة كيف حالك شو اخبارك وانها الغابة زمان عندك لاش مابتبين \end{arabtext}} \\ 
Whisper (EGY) & \RaggedRight{\begin{arabtext} \small يا زلمة كاف حالك شو اخبارك وانها الغابة اذا ما عنك ليش ما بتبين \end{arabtext}} \\

\bottomrule
\end{tabular}
}
\caption{\label{tab:examples-appendix}
    Outputs produced by \toolname~ when input is Egyptian and Jordanian. For Jordanian dialect, we do not have a finetuned model and Whisper (0-shot) performs best. Hence highlighting the lack of generalisation for various finetuned models to unseen dialects. 
}
\end{table*}